\documentclass[10pt,journal,compsoc]{IEEEtran}
\usepackage{graphicx,multirow} % Add all your packages here
\usepackage{algorithm}
\usepackage{algpseudocode}
\usepackage{caption}
\usepackage{longtable}
\usepackage{amsmath}
\usepackage{amssymb}
\usepackage{bm}
\usepackage{pstricks-add}
\usepackage{subfigure}
\usepackage{colortbl}
\DeclareGraphicsExtensions{.tif,.png,.jpg,.eps,.fig}
\graphicspath{{images_eps/}}
\usepackage{epstopdf}
\usepackage{fancyhdr} % 添加页眉页脚
\definecolor{Blue}{rgb}{1.0,0.75,0.8}
\ifCLASSOPTIONcompsoc
  \usepackage[nocompress]{cite}
\else
  \usepackage{cite}
\fi

\ifCLASSINFOpdf

\else

\fi

\begin{document}
\title{A Parallel Divide-and-Conquer based Evolutionary Algorithm for Large-scale Optimization}

\author{Peng~Yang,~\IEEEmembership{Member,~IEEE,}
    Ke~Tang,~\IEEEmembership{Senior Member,~IEEE,}
        and~Xin~Yao,~\IEEEmembership{Fellow,~IEEE}% <-this % stops a space
\IEEEcompsocitemizethanks{\IEEEcompsocthanksitem The authors are with Shenzhen Key Laboratory of Computational Intelligence, University Key Laboratory of Evolving Intelligent Systems of Guangdong Province，Department of Computer Science and Engineering, Southern University of Science and Technology, Shenzhen 518055, China. (e-mail: yangp@sustc.edu.cn; tangk3@sustc.edu.cn; xiny@sustc.edu.cn). Corresponding Author: Xin Yao.\protect\\
}% <-this % stops an unwanted space
\thanks{}}

\markboth{Submitted to IEEE Transactions on Parallel and Distributed Systems}%
{Shell \MakeLowercase{\textit{et al.}}: Bare Demo of IEEEtran.cls for Computer Society Journals}

\IEEEtitleabstractindextext{%
\begin{abstract}
Large-scale optimization problems that involve thousands of decision variables have extensively arisen from various industrial areas. 
As a powerful optimization tool for many real-world applications, evolutionary algorithms (EAs) fail to solve the emerging large-scale problems both effectively and efficiently. 
In this paper, we propose a novel Divide-and-Conquer (DC) based EA that can not only produce high-quality solution by solving sub-problems separately, but also highly utilizes the power of parallel computing by solving the sub-problems simultaneously. 
Existing DC-based EAs that were deemed to enjoy the same advantages of the proposed algorithm, are shown to be practically incompatible with the parallel computing scheme, unless some trade-offs are made by compromising the solution quality.
\end{abstract}

}

\maketitle

\IEEEdisplaynontitleabstractindextext

% \fi
\IEEEpeerreviewmaketitle

\IEEEraisesectionheading{\section{Introduction}\label{sec:introduction}}

\IEEEPARstart{N}{owadays}, with the trend of globalization, traditional decision-making components appear to produce only local optima. 
For example, the design of the global supply chain involves network planning, manufacturing production, warehousing operations, transportation planning, urban delivering, and so on \cite{Meixell2005global}. 
These components are heavily interdependent in nature, optimizing one may result in the degeneration of others. 
Only regarding them as an integrated optimization problem, could one produce a solution satisfying all components. 
In exchange, one has to optimize much more decision variables than usual, which brings in the widely-existed large-scale optimization problem \cite{Mei2018LSM,Di2018Exploring,Cui2018mSNP}.

Evolutionary Algorithms (EAs), which work by searching the solution space of the targeted problem iteratively and in a randomized way, have shown powerful performance in solving many real-world optimization problems \cite{Santander2015Parallel,Sheikh2016Evolutionary,Guo2015PSO,Zhu2018Eov,Li2018optimization}. 
Unfortunately, the search-based core makes EAs ineffective and inefficient for solving large-scale optimization problems for two reasons:
\begin{itemize}
\item[1)] As the number of decision variables increases, the solution space of the problem enlarges exponentially, preventing EAs exploring effectively within reasonable amount of search iterations.
\item[2)] As the search operators are applied to high-dimensional candidate solutions, the computational time costed at each iteration can become expensive. 
\end{itemize}
For many large-scale real-world optimization problems with real-time constraints, e.g., the placement of virtual machines in large-scale cluster \cite{Gulati2012vmware}, the above two drawbacks make EAs ill-equipped to handle them satisfactorily.

For the last decade, researchers have made great efforts on solving the above two drawbacks \cite{Santander2018Comparative,zhan2017cloudde,gong2015distributed}. 
Among the existing methods, the idea of Divide-and-Conquer (DC) has attracted most research attentions, as it is often viewed as an integrated solution for both drawbacks \cite{jia2018distributed}. 
In brief, DC-based EAs first decompose the original problem into multiple small-scale sub-problems, and then solve them respectively by existing EAs. 
Since small-scale sub-problems are usually easier to solve, significant reduction of the search iterations can be expected, which in turn leads to a satisfactory improvement of solution quality if the number of total iterations is fixed. 
Besides, ideally, multiple sub-problems can be solved in parallel, which can decrease the computational costs at each iteration dramatically. 

However, despite that existing DC-based EAs have remarkably reduced the search iterations \cite{yang2018turning,yang2008large,omidvar2014cooperative}, most of them cannot be directly implemented in parallel. 
This situation has been greatly neglected by researchers, because it is commonly believed that their serial workflow can be easily modified into parallel. 
In this paper, we first discuss that the modification of existing DC-based EAs into parallel could considerably compromise their effectiveness of reducing the search iterations, making it costly to accelerate the optimization via parallelization. 
Hence, a DC-based EA that is essentially parallelizable would be more meaningful and thus should be investigated. 

In fact, the difficulty of parallelizing existing DC-based EAs lies in the construction of objective functions for sub-problems. 
Specifically, to build the objective function for a sub-problem, existing works require the best partial solutions from other sub-problems. 
On the other hand, to decide such a best partial solution to a sub-problem, existing works require knowing its objective function first. 
This dining philosophers alike difficulty makes the sub-problems unable to be solved in parallel. 
In this paper, we propose a novel DC-based EA that is naturally suited to solve sub-problems in parallel. 
The core idea is to pre-select the best partial solution for each sub-problem via some subproblem-independent meta-model rather than the objective function of that sub-problem. 
By this means, the difficulty encountered when parallelizing existing DC-based EAs can be fully avoided.

The reminder of this paper is as follow. 
Section II reviews the background of this work. 
The difficulties of parallelizing existing DC-based EAs are analyzed in Section III. 
Inspired by that, a novel algorithm that is naturally suited to solve sub-problems in parallel is proposed Section IV. 
The above studies are empirically verified in Section V, where both the advantages and disadvantages of the proposed algorithm are discussed. 
Section VI concludes this paper and gives some directions for the future.

\section{Background}
\subsection{The Difficulties of EAs in Large-scale Optimization}
Given a real-world optimization problem, one can design an objective function either mathematically or by simulation. 
Without the loss of generality, let the minimization problem be an example. 
Generally, the optimization problem can be formulated as $\textbf{x}^*=\min_{\forall \textbf{x} \in \mathcal{X}} f(\textbf{x})$, where $\textbf{x}=[x_1,x_2,...,x_D]$ represents a $D$-dimensional candidate solution to $f$, $\mathcal{X}$ is the bounded $D$-dimensional solution space that contains all candidate solutions, and  $\textbf{x}^*$ is the optimum to $f$ within $\mathcal{X}$.

To solve an optimization problem, EAs actually search $\mathcal{X}$ for $\textbf{x}^*$ in an iterative way. 
Specifically, EAs work by first randomly initializing a population of (multiple) candidate solutions in $\mathcal{X}$, and then improving them iteratively during the search course, lastly outputting the best ever found candidate as the final solution. 
In the $t$-th iteration ($t<T$), a new population $\tilde{\textbf{X}}^t$ will be first generated based on the last population $\textbf{X}^t$ via some specific randomized search operators, and then the candidate solutions $\tilde{\textbf{x}}^t$ in the new population $\tilde{\textbf{X}}^t$ will be evaluated with $f$. 
After that, the candidate solutions in both $\textbf{X}^t$ and $\tilde{\textbf{X}}^t$ will be selected to form $\textbf{X}^{t+1}$ for the next iteration, in terms of their evaluated function values. 

When solving real-world optimization problems, the computational efficiency is an important requirement aside the solution quality, and sometimes can be a hard constraint. 
That is, the problem must be solved in a given time budget $T$, which would otherwise fail the quality of service provided by an application \cite{zhan2015cloud}. 
%For example, in the cloud data center, it is of importance to keep the workload in balance among hosts. Otherwise, the performance of virtual machines on some hosts would be severely deteriorated. 
%When such an imbalance occurs, one has to plan for re-scheduling the corresponding virtual machines over the available hosts in seconds \cite{zhan2015cloud}. 
In general, the computational efficiency of an EA is governed by two independent factors: the iterations of the whole search process and the computational time costed in each iteration. 
Unfortunately, both factors are very sensitive to the dimensionality $D$ of the problem. 
When $D$ becomes large, the real-time requirement of the problem will impose great challenges on EAs. 
On the other hand, these large-scale problems are ubiquitous \cite{zhang2014fuxi}. 
%For instance, in large data center, like Alibaba Cloud Aspara 5K cluster, there are usually thousands of hosts, on each of which runs multiple virtual machines \cite{zhang2014fuxi}. 
%In this case, in the problem of either re-scheduling or initially placing the virtual machines among the cluster, $D$ can easily increase to over 1000. 
Therefore, how to solve large-scale optimization problems with good enough solutions while keeping the computational costs sufficiently low lies in the core of the research of EAs.

\subsection{The Divide-and-Conquer based EAs}
In the literature, various ideas have been investigated for EAs on large-scale optimization problems. 
For reducing the search iterations, there are two major ways. 
One is to enhance the search ability of existing EAs by re-scheduling the local search operators \cite{molina2010ma,cheng2015competitive}; while the other is to simplify the problem via Divide-and-Conquer (DC) \cite{mei2016competitive,liu2001scaling,tang2017scalable} or Dimension Reduction \cite{kaban2016toward,qian2016scaling}. 
For saving the computational time in each iteration, parallel computing or distributed computing techniques are frequently employed to optimize individual solutions or decision variables on different threads \cite{zhan2017cloudde, gong2015distributed,alba1999survey,subbu2004modeling,qian2018distributed}. 
Among them, the DC methodology has been frequently introduced into EAs for large-scale optimization problems, since it can be regarded as an integrated solution to improve the above two factors for EAs.

The DC-based EAs consist of three major steps. First, the original $D$-dimensional problem is exclusively divided into $M$ $D$-dimensional sub-problems, where $\sum^M_{i=1} d_i=D$. 
Second, each sub-problem is solved by an EA, respectively. Last, the $d_i$-dimensional partial solution to each sub-problem is merged to form the $D$-dimensional complete solution to the original problem as the final output. 
Ideally, in case the sub-problems can be made independent of each other, they can be solved separately and simultaneously. 
By `separately', we mean that each EA only has to solve a $d_i$-dimensional small-scale sub-problem with a sub-space sized of $|\mathcal{X}|^\frac{d_i}{D}$, where $|\mathcal{X}|$ indicates the hyper-volume of the solution space $\mathcal{X}$. 
As a result, the iterations of the whole search process can be considerably reduced by DC, as $\sum^M_{i=1} |\mathcal{X}|^\frac{d_i}{D} < |\mathcal{X}|$. 
By `simultaneously', we mean that the $M$ sub-problems can be parallel solved with the aid of multiple processors of a work-station, distributed computing resources in traditional on premise data center, or cloud computing services, where the computational cost in each iteration can be made at most $M$ times cheaper. 
In case $M$ can be made close to $D$, the computational cost in each iteration enlarged by the increase of $D$ can be marginal.

In practice, the interdependencies among sub-problems are the main barrier that hinders the above two advantages of DC-based EAs, motivating rich volume of research efforts for eliminating them via advanced decomposition strategies \cite{Mahdavi2015metaheuristics}. 
Yang et al. \cite{yang2008large} proposed a random grouping technique that periodically randomly decomposed the decision variables into equal sized groups, showing that the probability of generating interdependent groups dropped down as the optimization process went on. 
Omdivar et al. \cite{omidvar2010more} and Yang et al. \cite{yang2008multilevel} later found that the group sizes and grouping frequency had great impacts on the performance of random grouping.
Chen et al. \cite{chen2010large} introduced the mathematical definition of interacting decision variables to guide the decomposition of sub-problems much more accurately. 
It checked for each pair of decision variables that whether the definition was violated. 
If so, such pair of decision variables were deemed to be interdependent and grouped together. 
More works \cite{sun2017recursive,mei2016competitive,omidvar2014cooperative} improved the grouping accuracy by referencing a tighter interdependency among decision variables, i.e., the additively separability \cite{gorman1968conditions}. 

Indeed, the above works have pushed the boundary of decomposition accuracy to a great extent, leading to significant reductions of search iterations on large-scale optimization problems. 
However, these works have not discussed how to parallel solving sub-problems in their paper. 
In fact, most existing works cannot be directly implemented in parallel. 
Though some other researchers tried to propose parallel schemes for existing DC-based EAs \cite{gong2015distributed,lu2014performance,wang2015cooperative,cao2017distributed}, in the next section, we discuss that those schemes will dramatically destroy the effectiveness of the above algorithms. 

\section{The Difficulties of Parallelizing existing DC-based EAs}
To explain the difficulties of existing DC-based EAs on parallelization, we first describe the phase of building objective functions for the sub-problems, which forms the key step of applying DC. 
Then we show that the mostly adopted way for building the objective functions actually follows a serial workflow that cannot be directly parallelized. 
Lastly, we analyze that, to parallelize existing DC-based EAs, the effectiveness of reducing search iterations could be considerably degenerated.

\subsection{Building Objective Functions for Sub-problems}
By applying DC-based EAs to a problem $f$, one has to first decompose $f$ into $M$ sub-problems, and then solve each of them by an EA. 
Before solving a sub-problem, its objective function should be known, so that its candidate partial solutions evaluated. 
Clearly, the original $D$-dimensional objective function $f$ cannot be directly used to evaluate the partial solutions to each $d_i$-dimensional sub-problem, due to the mismatch of dimensionality. 
Thus, one has to derive $M$ new $d_i$-dimensional objective functions from $f$ for the sub-problems, denoted as $f_i$, respectively. 
For black-box optimization problems, we may not know the explicit formula of $f$, one practical way to build $f_i$ is by complementing. 
Specifically, given a $d_i$-dimensional sub-problem, one can transform the $D$-dimensional $f$ into a $d_i$-dimensional $f_i$ by fixing the rest $(D-d_i)$-dimensional decision variables involved in $f$ with some feasible values. 
For simplicity, let us denote the set of decision variables belonging to the $i$-th sub-problem as $\mathcal{S}_i$, where $|\mathcal{S}_i|=d_i$, $i=1,2,...,M$, then we have

\begin{equation}
  f_i(\textbf{x}_{\mathcal{S}_i})=f([\textbf{v}_{\mathcal{S}_1}, \textbf{v}_{\mathcal{S}_2}, ..., \textbf{v}_{\mathcal{S}_{i-1}}, \textbf{x}_{\mathcal{S}_i}, \textbf{v}_{\mathcal{S}_{i+1}}, ..., \textbf{v}_{\mathcal{S}_M}]),
\end{equation} 

\noindent where $\textbf{v}_{\mathcal{S}_i}$ denotes the fixed values for the decision variables in the $j$-th sub-problem, $j=1,2,...,M$, and $j \neq i$. 
This is identical to first complement a $d_i$-dimensional partial solution into $D$-dimensional with those feasible values, and then evaluate it by $f$.

\subsection{The Serial Workflow of Existing DC-based EAs}
For each $i$-th sub-problem, there are in total $|\mathcal{X}|^{D-d_i}$ possibilities to fix the $(D-d_i)$-dimensional decision variables, resulting in $|\mathcal{X}|^{D-d_i}$ candidate $f_i$. 
As $f_i$ largely determines the search direction of an EA in the $i$-th sub-problem, setting different $f_i$ will lead to different search trajectories. 
Eventually, it will result in different qualities of the output solution within a given time budget, e.g., number of iterations. 

To achieve good results, existing DC-based EAs have tried different ways to build $f_i$ \cite{jansen2004cooperative,panait2010theoretical,panait2006archive,panait2005time,Tang2016adaptive,xiao2007self}. 
In general, the values of the $(D-d_i)$-dimensional decision variables are usually fixed by filling with dedicated partial solutions from the existing (sampled) population of other $M-1$ sub-problems. 
In other words, to build $f_i$ using Eq.(1), any $\textbf{v}_{\mathcal{S}_j}$ is a $d_j$-dimensional partial solution selected from the population that has already been generated in the $j$-th sub-problem. 
The difference among those related works lies in how the fixed values are selected. 
For example, Potter and De Jong \cite{potter1994cooperative} randomly selects the partial solutions in the current population (iteration) of other $M-1$ sub-problems. 
The algorithms in \cite{mei2016competitive,liu2001scaling,yang2008multilevel,chen2010large,jia2018distributed,omidvar2014cooperative} select the best partial solutions in the current population of other $M-1$ sub-problems. 
Other works like \cite{panait2005time,panait2006archive,jia2018distributed} select the best partial solutions from historical populations of other $M-1$ sub-problems. 
Among them, the $f_i$ built with the best partial solutions in the current population of other $M-1$ sub-problems has been empirically shown to perform the best, and thus has been mostly adopted in nowadays DC-based EAs. 

Specifically, let us denote the best partial solution in the $t$-th population of the $j$-th sub-problem as $\textbf{b}^t_{\mathcal{S}_j}$. 
Then, the mostly adopted objective function for the $i$-th sub-problem can be described as Eq.(2). 

\begin{equation}
  f_i^t(\textbf{x}^t_{\mathcal{S}_i})=f([\textbf{b}^t_{\mathcal{S}_1}, \textbf{b}^t_{\mathcal{S}_2}, ..., \textbf{b}^t_{\mathcal{S}_{i-1}}, \textbf{x}^t_{\mathcal{S}_i}, \textbf{b}^{t-1}_{\mathcal{S}_{i+1}}, ..., \textbf{b}^{t-1}_{\mathcal{S}_M}]).
\end{equation} 

We specially denote the objective function with a superscript $t$ as $f^t_i$ in Eq.(2), because it changes over iterations once some $\textbf{b}^t_{\mathcal{S}_j}$ varies from iteration to iteration, which is highly likely. 
Notice that, for the sub-problems indexed from the first to the $(i-1)$-th, their best partial solutions are selected from the $t$-th population. 
This scheme actually prevents the existing works from being directly parallelizable. 
That is, to build $f^t_{i+1}$, it requires $\textbf{b}^t_{\mathcal{S}_i}$ to be known. 
On the other hand, to obtain each $\textbf{b}^t_{\mathcal{S}_i}$, one has to evaluate all the candidate partial solutions in the $i$-th sub-problem via the objective function $f^t_i$. 
In a word, $f^t_i$ should be built before $f^t_{i+1}$. 
As a result, for any pair of sub-problems, they cannot be directly solved in parallel. 
For illustration, the workflow of Eq.(2) are shown in Fig.1(a).

%\begin{figure}[tbp]\renewcommand{\captionfont}{\footnotesize}
 % \centering 
 % \subfigure[Existing serial DC]{ 
 %   \label{fig:subfig:1_base} %% label for first subfigure 
 %   \includegraphics[width=0.35\linewidth]{Fig_1_(a).eps}} 
 % \subfigure[Existing parallel DC]{ 
 %   \label{fig:subfig:2_base} %% label for second subfigure 
 %   \includegraphics[width=0.35\linewidth]{Fig_1_(b).eps}} 
 % \subfigure[The ideal DC]{ 
 %   \label{fig:subfig:3_base} %% label for first subfigure 
 %   \includegraphics[width=0.35\linewidth]{Fig_1_(c).eps}}
 % \subfigure[The proposed DC]{ 
 %   \label{fig:subfig:4_base} %% label for second subfigure 
 %   \includegraphics[width=0.35\linewidth]{Fig_1_(d).eps}} 
 % \caption{The illustrations of different workflows of DC-based EAs. 
 % Among them, (a) shows the workflow of existing serial DC-based EAs, (b) shows the workflow of existing parallel DC-based EAs, (c) shows the workflow of the ideal way of building consistent sub-problem objectives, and (d) shows the workflow of the proposed algorithm. 
 % For simplicity, we only depict two sub-problems for each workflow. 
 % In each workflow, a black arrow from “a” to “b” means to obtain “b”, “a” should be known first.} 
%\end{figure}

\subsection{The Gaps between Serial and Parallel Workflows}
One can alternatively parallelize existing DC-based EAs by building $f^t_i$ with historical best partial solutions \cite{jia2018distributed}, in which all the $f^t_i$ can be built at the same time. 
A typical implementation of this idea can be described as Eq.(3),

\begin{equation}
  f_i^t(\textbf{x}^t_{\mathcal{S}_i})=f([\textbf{b}^{t-1}_{\mathcal{S}_1}, \textbf{b}^{t-1}_{\mathcal{S}_2}, ..., \textbf{b}^{t-1}_{\mathcal{S}_{i-1}}, \textbf{x}^t_{\mathcal{S}_i}, \textbf{b}^{t-1}_{\mathcal{S}_{i+1}}, ..., \textbf{b}^{t-1}_{\mathcal{S}_M}]),
\end{equation} 

\noindent where all the fixed values are selected as the best partial solutions from the $(t-1)$-th population \cite{jia2018distributed}. 
An illustration of the workflow of Eq.(3) is given in Fig.1(b). 
Because Eq.(3) looks quite close to Eq.(2), researchers often assume that the existing DC-based EAs can easily be parallelized with neglectable compromises on effectiveness \cite{jia2018distributed,Mahdavi2015metaheuristics,gong2015distributed}. 
However, in the following, we discuss that using Eq.(3) can lead to much worse optimization results than using Eq.(2), interring that it is difficult to implement the existing DC-based EAs in parallel.

Recall that, the final output of a DC-based EA is the merge of the best partial solutions to each sub-problem at the final $T$-th population. 
Also note that, any partial solution $\textbf{x}^T_{\mathcal{S}_i}$ with good quality to $f^T_i$ may not necessarily be a component of a good complete solution to $f$. 
This divergence stems from the complementary of $\textbf{x}^T_{\mathcal{S}_i}$ while being evaluated. 
To calculate the final solution quality with $f$, $\textbf{x}^T_{\mathcal{S}_i}$ is complemented with $\textbf{b}^T_{\mathcal{S}_j}$ in each $j$-th sub-problem, since the best partial solutions will be merged at the $T$-th population. 
However, neither Eq.(2) nor Eq.(3) can guarantee the same complementary to $\textbf{x}^T_{\mathcal{S}_i}$ while building $f^T_i$. 
That is, in some sub-problems, if not all, the best partial solutions in the $(T-1)$-th population will be selected, which are not necessarily to be equal to $\textbf{b}^T_{\mathcal{S}_j}$. 
The situation remains the same for any $t$-th ($t<T$) intermediate population. 
Therefore, the divergence between each $f^t_i$ and $f$ is accumulated along the whole search process, leading the existing DC-based EAs to possibly produce low-quality solution to $f$. 

Based on the discussion above, we can analyze the gap between Eq.(2) and Eq.(3), i.e., the serial and parallel workflows of existing DC-based EAs, as follows. 
We first describe the above-mentioned ideal way of building each $f^t_i$ as Eq.(4), which is consistent with the evaluation of the final solution quality to $f$.

\begin{equation}
  f_i^t(\textbf{x}^t_{\mathcal{S}_i})=f([\textbf{b}^t_{\mathcal{S}_1}, \textbf{b}^t_{\mathcal{S}_2}, ..., \textbf{b}^t_{\mathcal{S}_{i-1}}, \textbf{x}^t_{\mathcal{S}_i}, \textbf{b}^t_{\mathcal{S}_{i+1}}, ..., \textbf{b}^t_{\mathcal{S}_M}]).
\end{equation} 

\noindent Then we can measure the divergence of Eq.(2) and Eq.(4), and the divergence of Eq.(3) and Eq.(4). 
Finally, we give the gap between the serial and parallel workflows as the difference between the two divergences. 
For illustration, the workflow of Eq.(4) is shown in Fig.1(c), which clearly shows a dining philosophers alike dilemma as mentioned previously.

The divergence is calculated as the probability that either Eq.(2) or Eq.(3) works differently from Eq.(4) at each $t$-th iteration. 
By `different', we mean that not all $\textbf{b}^{t-1}_{\mathcal{S}_j}$ in Eq.(2) or Eq.(3) will be remained to the $t$-th population, i.e., $\textbf{b}^{t-1}_{\mathcal{S}_j} \neq \textbf{b}^t_{\mathcal{S}_j}$. 
For simplicity, let us assume that each sub-problem is equally important to the original problem, and the search operator employed in each sub-problem independently has a probability $1-p$ to produce a new better partial solution, i.e., $\textbf{b}^{t-1}_{\mathcal{S}_i} \neq \textbf{b}^t_{\mathcal{S}_i}$, $i=1,2,...,M$. 
Thus, with a probability $p$, we have $\textbf{b}^{t-1}_{\mathcal{S}_i} = \textbf{b}^t_{\mathcal{S}_i}$ for each $i$-th sub-problem.
Let us also denote the divergence of Eq.(2) and Eq.(4) over all sub-problems as $div_1$, and the divergence of Eq.(3) and Eq.(4) over all sub-problems as $div_2$. 
Then we have the following equations.

\begin{equation}
\begin{aligned}[b]
  div_1 &=1-\prod^M_{i=1} P(\textbf{b}^{t-1}_{\mathcal{S}_M}=\textbf{b}^t_{\mathcal{S}_M},...,\textbf{b}^{t-1}_{\mathcal{S}_i}=\textbf{b}^t_{\mathcal{S}_i})                \\
  &= 1-\prod^M_{i=1}\prod^M_{j=i}P(\textbf{b}^{t-1}_{\mathcal{S}_j}=\textbf{b}^t_{\mathcal{S}_j})    \\
  &= 1-\prod^M_{i=1}p^{M-i+1}    \\
  &= 1-p^{\frac{M(M-1)}{2}}
\end{aligned}              
\end{equation}

\begin{equation}
\begin{aligned}[b]
  div_2 &=1-\prod^M_{i=1} P(\textbf{b}^{t-1}_{\mathcal{S}_M}=\textbf{b}^t_{\mathcal{S}_M},...,\textbf{b}^{t-1}_{\mathcal{S}_{i+1}}=\textbf{b}^t_{\mathcal{S}_{i+1}}, \\
  &\qquad\qquad\,\,\,\,\,\,\,\,\,\, \textbf{b}^{t-1}_{\mathcal{S}_{i-1}}=\textbf{b}^t_{\mathcal{S}_{i-1}},...,\textbf{b}^{t-1}_{\mathcal{S}_1}=\textbf{b}^t_{\mathcal{S}_1})                \\
  &= 1-\prod^M_{i=1}\prod^M_{j=1,j \neq i}P(\textbf{b}^{t-1}_{\mathcal{S}_j}=\textbf{b}^t_{\mathcal{S}_j})    \\
  &= 1-\prod^M_{i=1}p^{M-1}    \\
  &= 1-p^{M(M-1)}
\end{aligned}              
\end{equation}

\noindent where the last term in either Eq.(5) or Eq.(6) describes the probability that all $\textbf{b}^{t-1}_{\mathcal{S}_j}$ in Eq.(2) or Eq.(3) are remained to the $t$-th population, i.e., $\textbf{b}^{t-1}_{\mathcal{S}_j}=\textbf{b}^t_{\mathcal{S}_j}$. 
Therefore, by building $f^t_i$ with Eq.(2), there is a probability $div_1$ that not all $\textbf{b}^t_{\mathcal{S}_j}$ is a good component to $f$. 
Similarly, by building $f^t_i$ with Eq.(3), there is a probability $div_2$ that not all $\textbf{b}^t_{\mathcal{S}_j}$ is a good component to $f$. 
Then the gap between Eq.(2) and Eq.(3) at each iteration can be expressed as $\frac{div_1}{div_2}=1+p^{\frac{M(M-1)}{2}}>1$, and the accumulated gap will exponentially explode as $(\frac{div_1}{div_2})^T=(1+p^{\frac{M(M-1)}{2}})^T \to \infty$, in case the total iterations $T \to \infty$. 
As a result, by parallelizing existing DC-based EAs with Eq.(3), the reduction of the search iterations, or the quality of the final output if $T$ is fixed, can be considerably deteriorated.

\section{The Proposed Parallel DC-based EA}

As discussed above, Eq.(4) can evaluate partial solutions with the quality consistent with $f$. 
Unfortunately, existing works cannot adopt it, due to a difficulty similar to the so-called dining philosophers. 
That is, in Eq.(4), to obtain each $\textbf{b}^t_{\mathcal{S}_j}$, one has to evaluate all the candidate partial solutions in the $j$-th sub-problem via the objective function $f^t_j$. 
On the other hand, to build each $f^t_i$, $i=1,2,...,M$ and $i \neq j$, one has to first obtain the best partial solutions $\textbf{b}^t_{\mathcal{S}_j}$ in other $j$-th sub-problems. 
In a word, $f^t_i$ and $f^t_j$ are mutually prior conditions required by each other. 
As a result, for any pair of sub-problems, their objective functions cannot be built simultaneously and thus themselves cannot be solved in parallel. 
For illustration, the workflow of Eq.(4) are shown in Fig.1(c).
 
In this paper, we propose not to decide each $\textbf{b}^t_{\mathcal{S}_j}$ by $f^t_j$. 
Instead, each $\textbf{b}^t_{\mathcal{S}_j}$ is first pre-selected via a meta-model, denoted as $g^t_j$, which is not built upon any $\textbf{b}^t_{\mathcal{S}_i}$. 
By this means, the dining philosophers alike difficulty of Eq.(4) is tackled, as shown in Fig.1(d). 
Based on that, the resultant algorithm, called Naturally Parallelizable Divide-and-Conquer (NPDC), is thus both naturally parallelizable and consistent with $f$. 
In this section, the meta-model $g^t_j$ is first detailed, then the proposed NPDC algorithm is described, lastly the parallelism of NPDC is discussed in details.
 
\subsection{The Proposed Meta-model of NPDC}
The meta-model is basically designed for the (1+1)-EA paradigm \cite{qian2017running,jansen2004cooperative}. 
The (1+1)-EA paradigm is defined as that an offspring solution will be first generated based its parent solution, and then directly compared to the parent for survival. 
Therefore, in the $j$-th sub-problem at the $t$-th population, $g^t_j$ works for pair-wise comparing one parent partial solution and its offspring partial solution, and deciding the better one. 
The underlying reason is two-folds. First, the (1+1)-EA paradigm is very simple, for which the meta-model will be kept computationally efficient and easily understandable. 
Second, the (1+1)-EA paradigm is widely existed in many well-established EAs, e.g., Differential Evolution (DE) \cite{storn1997differential}, Particle Swarm Optimization (PSO) \cite{shi1998modified}, for which the offspring and its parent has the one-on-one relationship. 
The meta-model can easily be generalized to those kinds of population-based EAs. 
To this end, one can simply employ $\lambda$ meta-models if the population size is $\lambda$, each of which works for an individual in the population under the (1+1)-EA paradigm. 
For other EAs like Estimation of Distribution Algorithms \cite{yang2015improving} that directly operate at the population level rather than the individual level, there does not exist the one-on-one relationship between parents and offsprings. 
Thus, the proposed meta-model cannot be directly applied to them. 

The meta-model is also designed for the 1-dimensional sub-problems. 
In the literature, there are lots of meta-models that aim to evaluate the quality of a candidate solution instead of the original computationally expensive objective function \cite{Tabatabaei2015,LiuBo2014,jin2011surrogate}. 
However, these meta-models often require huge number of labelled data since they usually simulate the global solution space with relatively high-dimension, which in turn require extensive calls of function evaluations. 
As there are $M \times \lambda$ meta-models in NPDC, directly employing existing meta-models in this paper is computationally unaffordable. 
Considering this, in order to keep it efficient enough, the proposed meta-model only focuses on estimating the local area of a 1-dimensional sub-space, where the number of labelled data can be minimized. 
In this regard, $f$ is actually decomposed into $D$ sub-problems, each of which is 1-dimensional, i.e., $M=D$.

For simplicity and clarity, we will illustrate the meta-model in case of $\lambda = 1$. 
To be specific, under the (1+1)-EA paradigm, the best partial solution at the last iteration can be simply regarded as the parent at the current iteration. 
Therefore, the parent partial solution in the $j$-th sub-problem at the $t$-th population can be denoted as $b^{t-1}_j$. 
Here we use the denotation $b^{t-1}_j$, rather than $\textbf{b}^{t-1}_{\mathcal{S}_j}$ as in the previous sections, because each $j$-th sub-problem is actually 1-dimensional and thus $\mathcal{S}_j=\{j\}$. 
Then, $g^t_j$ works for pair-wise comparing $b^{t-1}_j$ and $\tilde{x}^t_j$, and deciding the better one as $b^t_j$, as Eq.(7) shows.

\begin{equation}
  b^t_j=g^t_j(b^{t-1}_j, \tilde{x}^t_j),
\end{equation} 

\noindent where $\tilde{x}^t_j$ is the offspring partial solution generated based on its parent $b^{t-1}_j$. 

As it is in 1-dimensional, $b^{t-1}_j$ and $\tilde{x}^t_j$ can be directly compared by their values. 
Therefore, the meta-model $g^t_j$ actually learns the relationship between their qualities and values. 
More specifically, the value of the offspring $\tilde{x}^t_j$ can be either larger or smaller to the value of its parent $b^{t-1}_j$, and the quality of $\tilde{x}^t_j$ can be either superior or inferior to the quality of $b^{t-1}_j$\footnote{The equivalent case is highly unlikely and thus can be practically assigned to either inequivalent case.}. 
If the landscape of the solution space changes smoothly, $g^t_j$ can use the historical comparison results to infer the current comparison between $b^{t-1}_j$ and $\tilde{x}^t_j$ in a probabilistic way. 
Note that, the local landscape is highly likely to be asymmetrical to $b^{t-1}_j$. 
Then we can use $PL^t_j$ to denote the probability that the value of $\tilde{x}^t_j$ is larger than the value of $b^{t-1}_j$ and the quality of $\tilde{x}^t_j$ is superior to the quality of $b^{t-1}_j$. 
Similarly, $PS^t_j$ denotes the probability that the value of $\tilde{x}^t_j$ is smaller than the value of $b^{t-1}_j$ and the quality of $\tilde{x}^t_j$ is superior to the quality of $b^{t-1}_j$. 
Based on that, $g^t_j$ can be defined as Eq.(8). 

\begin{eqnarray} 
g^t_j(b^{t-1}_j, \tilde{x}^t_j)  = \left\{ 
\begin{array}{ll}  
 \tilde{x}^t_j & \textbf{if} \quad \tilde{x}^t_j < b^{t-1}_j \,\, \textbf{and} \,\, PS^t_j < r\\ 
 & \textbf{or} \,\,\,\, \tilde{x}^t_j > b^{t-1}_j \,\, \textbf{and} \,\, PL^t_j < r \\
 b^{t-1}_j &  \textbf{otherwise} 
\end{array} 
\right. 
\end{eqnarray}

\noindent where $r$ indicates a function that returns a random variable uniformly sampled in the range of $[0,1]$. 
Eq.(8) actually says that given $\tilde{x}^t_j$ is larger/smaller than $b^{t-1}_j$, there is a probability $PL^t_j$/$PS^t_j$ that $\tilde{x}^t_j$ is superior to $b^{t-1}_j$. 
Ideally, if the landscape increases/decreases monotonously along the 1-dimensional solution space, $PL^t_j$/$PS^t_j$ equals to 1.00. 
Unfortunately, it is often the case that the landscape changes from areas to areas in the solution space. 
Thus, both $PL^t_j$ and $PS^t_j$ need to be tuned over iterations. 
Assume that the landscape changes smoothly and the value of $\tilde{x}^t_j$ can be generated close to $b^{t-1}_j$, both probabilities can be adjusted slightly over iterations according to the 1/5 successful rule \cite{hansen2015evolution}, which is a heuristic rule for tuning parameters locally, as Eq.(9). 

\begin{equation} 
\begin{split}
&PS^{t+1}_j = PS^t_j \cdot \textrm{exp}^{\frac{1}{\sqrt{2}}}[\mathbb{I}_{\tilde{x}^t_j< b^{t-1}_j}\cdot(\mathbb{I}_\Theta-\frac{1}{5})] \\
&PL^{t+1}_j = PL^t_j \cdot \textrm{exp}^{\frac{1}{\sqrt{2}}}[\mathbb{I}_{\tilde{x}^t_j> b^{t-1}_j}\cdot(\mathbb{I}_\Theta-\frac{1}{5})] \\
\textrm{where} \,\, \Theta &= f([b^t_1,b^t_2,...,b^t_D]) < f([b^{t-1}_1,b^{t-1}_2,...,b^{t-1}_D]).
\end{split}
\end{equation}

In Eq.(9), $\mathbb{I}_\Theta$ is an indicator function that returns 1 if event $\Theta$ is true and 0 otherwise. 
$\Theta$ denotes the comparison on the qualities between each pair of $b^{t-1}_j$ and $b^t_j$. 
It is calculated after the $b^t_j$ in each $j$-th sub-problem is decided via $g^t_j$.
It says that if the quality of $\tilde{x}^t_j$ is superior to the quality of $b^{t-1}_j$, and if the value of $\tilde{x}^t_j$ is larger/smaller than the value of $b^{t-1}_j$, the probability $PL^t_j$/$PS^t_j$ should be enlarged with a factor of $\textrm{exp}^{\frac{1}{\sqrt{2}}}(\frac{4}{5})$. 
Otherwise, if the quality of $\tilde{x}^t_j$ is inferior to the quality of $b^{t-1}_j$, the corresponding probability should be reduced with a factor of $\textrm{exp}^{\frac{1}{\sqrt{2}}}(-\frac{1}{5})$. 
After tuning the two probabilities with Eq.(9), $g^t_j$ is actually adjusted for one iteration and thus re-denoted as $g^{t+1}_j$. 

Both $PS^0_j$ and $PL^0_j$ are initialized as 1.00 at the beginning of the search. 
This means that the randomly initialized parent $b^0_j$ is assumed to be always inferior to its offspring $\tilde{x}^1_j$, which is reasonable. 
During the adjustment, both $PS^0_j$ and $PL^0_j$ should be kept always no larger than 1.00, and no smaller than 0.00. 
However, if either of them equals 0.00, there will be no chance for them to be changed according to Eq.(9). 
And the search process will be stagnated since any $\tilde{x}^t_j$ will be regarded as inferior to $b^{t-1}_j$ and thus $b^t_j$ always equals to $b^{t-1}_j$ according to Eq.(9). 
In practice, we let them to be no smaller than a very small value, i.e., $\frac{2}{D}$. 
Given an offspring has averagely 50\% probability to be generated either smaller or larger than the parent, then it is guaranteed that each decision variable has at least $\frac{1}{D}$ probability to be changed in each iteration. 

\subsection{The Details of NPDC}
NPDC follows the divide-and-conquer framework. 
Specifically, NPDC firstly decomposes the original objective function $f$ into $D$ sub-problems, each of which exclusively contains one decision variable. 
Then each sub-problem is solved in parallel. At the $T$-th iteration, the best-ever found partial solution $b^T_j$ to each $j$-th sub-problem is merged to form the final solution $\textbf{b}^T$ to the original problem $f$. 

\begin{algorithm}[t] 
   \caption{NPDC($f$, $T, \lambda, n$)}  
   \begin{algorithmic}[1] %每行显示行号 
       \State Divide $f$ into $D$ exclusive sub-problems.
       \State \textbf{For} $i = 1$ \textbf{to} $\lambda$ 
       \State \quad \textbf{For} $j = 1$ \textbf{to} $D$ 
       \State \quad\quad $PS^1_{i,j}=1.0$; $PL^1_{i,j}=1.0$; and $\sigma^t_{i,j}=1.0$.
       \State \quad\quad Initialize $x^1_{i,j}$ uniformly randomly; Let $b^0_{i,j}=x^1_{i,j}$.
       %\State \quad \textbf{EndFor}
       \State \quad $FB_i=f([b^0_{i,1},b^0_{i,2},...,b^0_{i,D}])$.
       %\State \textbf{EndFor} 
       \State \textbf{For} $t=1$ \textbf{to} $\frac{T}{\lambda}$
           \State \quad \textbf{For} $i = 1$ \textbf{to} $\lambda$  
           \State \quad\quad  \textbf{For} $j = 1$ \textbf{to} $D$ 
           \State \quad\quad\quad  \textbf{If} $r>0.5$
           \State \quad\quad\quad\quad  $\tilde{x}^t_j=b^{t-1}_j+\sigma^t_j\cdot\mathcal{N}(0,1)$.
           \State \quad\quad\quad  \textbf{Else} 
           \State \quad\quad\quad\quad  $\tilde{x}^t_j=b^{t-1}_j+\sigma^t_j\cdot\mathcal{C}(0,1)$.
          % \State \quad\quad\quad  \textbf{EndIf}
           \State \quad\quad\quad  $b^t_j=g^t_j(b^{t-1}_j, \tilde{x}^t_j)$.
         %  \State \quad\quad  \textbf{EndFor}
           \State \quad\quad  $\widetilde{FB_i}=f([b^t_{i,1},b^t_{i,2},...,b^t_{i,D}])$.
           \State \quad\quad  \textbf{For} $j = 1$ \textbf{to} $D$ 
           \State \quad\quad\quad  $PS^{t+1}_j = PS^t_j \cdot \textrm{exp}^{\frac{1}{\sqrt{2}}}[\mathbb{I}_{\tilde{x}^t_j< b^{t-1}_j}\cdot(\mathbb{I}_{\widetilde{FB_i}<FB_i}-\frac{1}{5})]$.
           \State \quad\quad\quad  $PL^{t+1}_j = PL^t_j \cdot \textrm{exp}^{\frac{1}{\sqrt{2}}}[\mathbb{I}_{\tilde{x}^t_j< b^{t-1}_j}\cdot(\mathbb{I}_{\widetilde{FB_i}<FB_i}-\frac{1}{5})]$.
           \State \quad\quad\quad  $\sigma^{t+1}_j = \sigma^t_j \cdot \textrm{exp}^{\frac{1}{\sqrt{2}}}[\mathbb{I}_{\tilde{x}^t_j< b^{t-1}_j}\cdot(\mathbb{I}_{\tilde{FB_i}<FB_i}-\frac{1}{5})]$.
        %   \State \quad\quad  \textbf{EndFor} 
           \State \quad\quad  \textbf{If} $\widetilde{FB_i}<FB_i$ 
           \State \quad\quad\quad  $FB_i=\widetilde{FB_i}$.
           \State \quad\quad  \textbf{Else}
           \State \quad\quad\quad  \textbf{For} $j = 1$ \textbf{to} $D$
           \State \quad\quad\quad\quad  $b^t_{i,j}=b^{t-1}_{i,j}$.
        %   \State \quad\quad  \textbf{EndFor} 
      %  \State \textbf{EndFor}  
       \State \textbf{Output} $\textbf{min}_{1 \leq i \leq \lambda}(FB_i)$.
   \end{algorithmic} 
\end{algorithm} 

In order to keep NPDC compatible with the proposed meta-model, the optimization of each sub-problem should comply with two considerations. 
First, each sub-problem should be solved in a (1+1)-EA paradigm. 
%That is, at the $t$-th iteration, the parent $b^{t-1}_j$ generates only one offspring $\tilde{x}^t_j$ with some search operator, and competes with $\tilde{x}^t_j$ directly for survival. 
Second, the offspring should be generated close to its parent, so that the historical comparison results can be helpful to the current comparison between them. 
Given this, the Gaussian mutation operator, as shown in Eq.(10), is well-suited to be the search operator in NPDC since it generates the offspring closer to the parent with higher probability.

\begin{equation}
  \tilde{x}^t_j=b^{t-1}_j+\sigma^t_j\cdot\mathcal{N}(0,1),
\end{equation} 

\noindent where $\sigma^t_j$ indicates the search step-size, and $\mathcal{N}(0,1)$ is a Gaussian random variable with zero mean and standard deviation 1. 
On the other hand, if we think from the perspective of the search effectiveness, the search operator should generate offsprings quite different from the parent, so that the search process can go faster and has higher probability to jump away from premature convergence. 
Such purpose can be realized by the Cauchy mutation operator \cite{yao1999evolutionary}, as shown in Eq.(11), since it has much `wider' probability distribution than the Gaussian mutation.

\begin{equation}
  \tilde{x}^t_j=b^{t-1}_j+\sigma^t_j\cdot\mathcal{C}(0,1),
\end{equation} 

\noindent where $\mathcal{C}(0,1)$ indicates a random variable subject to the standard Cauchy distribution. 

In order to make a good balance between the effectiveness of search process and that of the meta-model, NPDC chooses either Eq.(10) or Eq.(11) to generate an offspring at random. 
Specifically, at each iteration, NPDC uniformly generates a random variable from [0,1]. 
If its value is larger than 0.5, the offspring is generated using Eq.(10), which would otherwise be generated using Eq.(11). 
Generally, the search step-size $\sigma^t_j$ in both Eq.(10) and Eq.(11) can be adapted during the search and may also vary over sub-problems. 
To make it simple, each $\sigma^t_j$ is initialized as 1.00, and adjusted at each iteration, according to the 1/5 successful rule again, as given in

\begin{equation}
  \sigma^{t+1}_j = \sigma^t_j \cdot \textrm{exp}^{\frac{1}{\sqrt{2}}}[\mathbb{I}_{\tilde{x}^t_j \neq b^{t-1}_j}\cdot(\mathbb{I}_\Theta-\frac{1}{5})].
\end{equation} 

\noindent Eq.(12) says that, as long as $\tilde{x}^t_j \neq b^{t-1}_j$, $\sigma^t_j$ will be adjusted. More specifically, if the quality of $b^t_j$ is superior to the quality of $\tilde{x}^t_j$, $\sigma^t_j$ will be enlarged with a factor of $\textrm{exp}^{\frac{1}{\sqrt{2}}}(\frac{4}{5})$. 
Otherwise, $\sigma^t_j$ will be reduced with a factor of $\textrm{exp}^{\frac{1}{\sqrt{2}}}(-\frac{1}{5})$. 
The qualities of $\tilde{x}^t_j$ and $b^{t-1}_j$ are compared using $\Theta$ in Eq.(9).

As a summary, the detailed pseudo-code of NPDC is presented in Algorithm 1 for illustration. 
In order to generalize the description to the population-based EAs composed of multiple (1+1)-EA paradigms, the denotation of each $b^t_j$ is modified to $b^t_{i,j}$, where $i=1,2,...,\lambda$ denotes the $i$-th (1+1)-EA paradigm, or say the $i$-th individual. 
Other denotations are also modified accordingly. 
As seen that, NPDC initializes the optimization process at steps 1-5 and records the best complete solution at step 6. 
Then for each sub-problem, NPDC employs a sub-problem optimizer consisting with $\lambda$ (1+1)-EA paradigm based search operators. 
The search operators generate offsprings at steps 10-13 with the Gaussian mutation and Cauchy mutation. 
Other EAs with the (1+1)-EA paradigm, e.g., PSO and DE, can easily be incorporated into NPDC by replacing steps 10-13 with their specific search operators. 
The offsprings and their parents are pairwise compared and selected to the next iteration at step 14 using the meta-model. 
The pre-selected best partial solutions in each sub-problem is merged and evaluated at step 15, where the original objective function $f$ will be called once for a (1+1)-EA paradigm at one iteration. 
In a word, at each iteration of the whole search, NPDC consumes $\lambda$ function evaluations at step 15. 
The meta-models and the search operators are adjusted at steps 16-19, based on the qualities of the merged solutions evaluated by $f$. 
The best complete solution of each (1+1)-EA paradigm is respectively updated at steps 20-24. 
When the whole optimization process is iterated for $\frac{T}{\lambda}$ times, i.e., NPDC consumes $T$ function evaluations, the best complete solution among all merges of the $\lambda$ (1+1)-EA paradigm is output at step 25.  

\subsection{The Parallelism of NPDC}
As can be seen from Algorithm 1, NPDC is parallelizable on two levels. 
First, NPDC is fully parallelizable on the individual level, since each of the $\lambda$ (1+1)-EA paradigms acts fully independently with others. 
Second, NPDC is parallelizable on the decision variables level. 
Specifically, only step 6 and step 15 require merging $b^t_{i,j}$ from each $j$-th sub-problem, while other steps can be processed independently. 
As discussed earlier, each $b^t_{i,j}$ required by step 6 and step 15 will not introduce any temporal interdependencies among sub-problems, since NPDC can pre-select $b^t_{i,j}$ by the meta-model of each sub-problem itself at step 14. 
Therefore, as long as the workload for solving each sub-problem is in balance, which is highly likely, each $b^t_{i,j}$ can be collected simultaneously from the sub-problems for function evaluations at step 6 and step 15.

Based on that, to implement NPDC in parallel, one can simply allocate the computational tasks of each sub-problem to a machine. 
If the number of available machines is smaller than $D$, one can either allocate an equal number of sub-problems to each machine or extend one iteration of Algorithm 1 to multi-round. 
When those slave machines have finished their jobs of sub-problem optimizers for one iteration, the data of $b^t_{i,j}$ will be transmitted to a master machine for executing step 6 and step 15. 
Then the information of the corresponding function evaluations will be transmitted from the master machine back to each slave machine. 
In this regard, the only bottleneck of parallelizing NPDC is the computational efficiency of the calls of $f$ at step 6 and step 15, which cannot be parallelized unless the explicit formula of $f$ can be known and mathematically decomposable. 

Specifically, the parallelism of NPDC can be quantified in terms of the speed-up \cite{Lucas2018Visibility}. 
The speed-up is a key indicator for measuring parallel efficiency \cite{vargas2017hybrid}. 
It measures the ratio of $T_1$ over $T_N$, i.e., the elapsed computational time of a task executed on 1 and $N$ processors. 
Let us denote the computational time costed by the $T$ function evaluations as $T_{FE}$, then given the above discussions we have $T_1-T_{FE}=N\cdot(T_N-T_{FE})$. 
As a result, the speed-up of NPDC is ideally given as Eq.(13).

\begin{equation}
\begin{aligned}[b]
  \frac{T_1}{T_N}=\frac{T_1}{T_{FE}+(T_N-T_{FE})} 
  &=\frac{T_1}{T_{FE}+\frac{1}{N}\cdot(T_1-T_{FE})} \\
  &=\frac{N}{1+\frac{T_{FE}}{T_{1}}\cdot(N-1)}.
\end{aligned}
\end{equation} 

\noindent If $\frac{T_{FE}}{T_1} \to 0$, the speed-up ratio converges to $N$, which means NPDC enjoys a linear speed-up that the total computational time can be reduced by $N$ times if $N$ processors are used. 
On the other hand, if $\frac{T_{FE}}{T_1} \to 1$, the speed-up ratio converges to 1, which means that parallelization does not make any sense to NPDC. 
Also note that, although we cannot distribute the computational task of each function evaluation to multiple processors, the burden of $T_{FE}$ can still be alleviated by distributing multiple function evaluations to multiple processors if a population consists of multiple individuals \cite{dubreuil2006analysis, wu2004incremental, zhan2017cloudde}.
 
\section{Empirical Studies}
This paper arises four questions. 
First, how differently the existing serial DC-based works behave from their parallel counterparts, in terms of the final solution quality? 
Second, does NPDC perform better than the existing DC-based works, in terms of the final solution quality? 
Third, how is the parallelism of NPDC in practice? Lastly, how does the proposed meta-model impact on the performance of NPDC.
In the empirical studies, four groups of comparisons are conducted to answer these questions, respectively. 

%\begin{figure*}[tbp]\renewcommand{\captionfont}{\footnotesize}
 % \centering 
 % \subfigure[Comparisons of DC-NG and DC-NG-P]{ 
 %   \label{fig:subfig:1_base} %% label for first subfigure 
  %  \includegraphics[width=0.32\linewidth]{Fig_2_(a).eps}} 
 % \subfigure[Comparisons of DC-RG and DC-RG-P]{ 
 %   \label{fig:subfig:2_base} %% label for second subfigure 
 %   \includegraphics[width=0.32\linewidth]{Fig_2_(b).eps}} 
 % \subfigure[Comparisons of DC-DG and DC-DG-P]{ 
 %   \label{fig:subfig:3_base} %% label for first subfigure 
 %   \includegraphics[width=0.32\linewidth]{Fig_2_(c).eps}}
 % \caption{The pairwise comparisons of the existing DC-based EAs and their parallel counterparts, in terms of the final solution quality.} 
%\end{figure*}

\subsection{Experiment protocol}
The CEC'2010 large-scale optimization benchmark covers different degrees of separability and multi-modality, which are the main difficulties for large-scale optimization problems \cite{tang2010benchmark}, and thus has been commonly used to verify the performance of large-scale optimization algorithms \cite{omidvar2017dg2,chen2010large,omidvar2014cooperative}. 
Given this, it is also adopted as the test suite in the empirical studies. 

All the 20 problems in the test suite involve 1000 decision variables, i.e., $D$=1000. 
The time budget, i.e., the total number of function evaluations of each run, is set to 3e6, which is widely used in the literature \cite{omidvar2017dg2,chen2010large,omidvar2014cooperative}. 
Each run of an algorithm terminates when the time budget runs out. 
The function error is used to measure the quality of the final solution, i.e., the difference between the objective function value of the final output solution and that of the optimal solution to the problem (which are known for all tested problems as 0.0). 
In a word, the smaller the function error is, the better the quality of the final solution will be. 
All the compared algorithms are repeated on each problem for 20 runs. 
The function errors of the corresponding solutions are recorded and averaged over 20 runs to represent the performance of algorithms on a problem. 
The two-sided Wilcoxon rank-sum test at a 0.05 significance level is also conducted to see whether the performances of pairwise compared algorithms are statistically significantly different. 
All experiments are conducted on a workstation with 2 CPUs specified with Intel(R) Xeon(R) CPU E5-2620 v4 @ 2.10 GHz (in total 16 physical cores), 128 GB memory size, 256 GB SSD, and Ubuntu 16.04 LTS.

{\renewcommand\baselinestretch{0.8}\selectfont
\begin{table}[tbp]\scriptsize
\centering  % 表居中
\caption{The average and standard deviation of function errors on the 1000-dimensional CEC'2010 large-scale global optimization benchmark.}

\begin{tabular}{ccccccc}
\hline
\multicolumn{2}{c}{\textbf{Algo.}} &\textbf{DC-NG} &\textbf{DC-RG} &\textbf{DC-DG} &\textbf{NPDC}\\ \hline
\hline
\multirow{2}*{\boldsymbol{$F_{1}$}} 
&Mean    &4.30e-23  &2.65e+05  &7.65e+04  &\cellcolor{lightgray}0.00e+00\\ 
&Std     &4.72e-23  &5.56e+04  &1.92e+04  &0.00e+00\\ \hline

\multirow{2}*{\boldsymbol{$F_{2}$}} 
&Mean    &\cellcolor{lightgray}1.67e+03  &7.63e+03  &2.67e+03  &8.38e+03\\ 
&Std     &1.61e+02  &6.20e+02  &4.57e+02  &3.69e+02\\ \hline

\multirow{2}*{\boldsymbol{$F_{3}$}} 
&Mean    &\cellcolor{lightgray}1.98e+01  &1.99e+01  &1.99e+01  &1.99e+01\\ 
&Std     &1.91e-02  &1.53e-02  &1.65e-02  &1.29e-02\\ \hline

\multirow{2}*{\boldsymbol{$F_{4}$}} 
&Mean    &1.55e+12  &4.16e+10  &8.82e+10  &\cellcolor{lightgray}1.66e+10\\ 
&Std     &7.30e+11  &1.56e+10  &2.46e+10  &9.79e+09\\ \hline

\multirow{2}*{\boldsymbol{$F_{5}$}} 
&Mean    &8.79e+08  &7.68e+08  &7.28e+08  &\cellcolor{lightgray}5.71e+08\\ 
&Std     &1.60e+08  &1.05e+08  &1.14e+08  &1.54e+08\\ \hline

\multirow{2}*{\boldsymbol{$F_{6}$}} 
&Mean    &1.98e+07  &1.98e+07  &1.98e+07  &\cellcolor{lightgray}1.98e+07\\ 
&Std     &9.39e+04  &6.17e+04  &8.96e+04  &7.16e+04\\ \hline

\multirow{2}*{\boldsymbol{$F_{7}$}} 
&Mean    &6.32e+06  &9.64e+04  &1.64e-17  &\cellcolor{lightgray}1.22e-18\\ 
&Std     &5.05e+06  &1.86e+04  &9.42e-18  &1.98e-19\\ \hline

\multirow{2}*{\boldsymbol{$F_{8}$}} 
&Mean    &1.95e+08  &2.01e+07  &2.00e+07  &\cellcolor{lightgray}1.20e+06\\ 
&Std     &2.65e+08  &3.25e+07  &4.70e+07  &1.87e+06\\ \hline

\multirow{2}*{\boldsymbol{$F_{9}$}} 
&Mean    &1.30e+07  &9.17e+06  &9.72e+06  &\cellcolor{lightgray}4.48e+06\\ 
&Std     &1.30e+06  &7.99e+05  &8.35e+05  &5.12e+05\\ \hline

\multirow{2}*{\boldsymbol{$F_{10}$}} 
&Mean    &8.84e+03  &1.26e+04  &\cellcolor{lightgray}8.60e+03  &1.23e+04\\ 
&Std     &4.49e+02  &6.12e+02  &4.80e+02  &4.11e+02\\ \hline

\multirow{2}*{\boldsymbol{$F_{11}$}} 
&Mean    &\cellcolor{lightgray}2.18e+02  &2.18e+02  &2.19e+02  &2.19e+02\\ 
&Std     &2.43e-01  &2.72e-01  &2.28e-01  &2.49e-01\\ \hline

\multirow{2}*{\boldsymbol{$F_{12}$}} 
&Mean    &5.23e+01  &2.73e-05  &7.14e-12  &\cellcolor{lightgray}3.89e-15\\ 
&Std     &3.91e+01  &2.35e-05  &2.75e-12  &2.78e-15\\ \hline

\multirow{2}*{\boldsymbol{$F_{13}$}} 
&Mean    &1.69e+03  &1.01e+03  &1.23e+03  &\cellcolor{lightgray}1.40e+01\\ 
&Std     &7.69e+02  &8.58e+02  &6.93e+02  &1.38e+01\\ \hline

\multirow{2}*{\boldsymbol{$F_{14}$}} 
&Mean    &2.84e+07  &2.41e+07  &3.19e+07  &\cellcolor{lightgray}1.79e+07\\ 
&Std     &1.50e+06  &1.79e+06  &2.75e+06  &1.35e+06\\ \hline

\multirow{2}*{\boldsymbol{$F_{15}$}} 
&Mean    &1.56e+04  &1.55e+04  &\cellcolor{lightgray}1.36e+04  &1.55e+04\\ 
&Std     &4.66e+02  &4.61e+02  &7.00e+02  &4.54e+02\\ \hline

\multirow{2}*{\boldsymbol{$F_{16}$}} 
&Mean    &\cellcolor{lightgray}3.97e+02  &3.97e+02  &3.98e+02  &3.98e+02\\ 
&Std     &2.57e-01  &3.21e-01  &4.29e-01  &3.94e-01\\ \hline

\multirow{2}*{\boldsymbol{$F_{17}$}} 
&Mean    &8.90e+01  &2.19e-01  &1.87e-04  &\cellcolor{lightgray}1.24e-06\\ 
&Std     &6.97e+01  &4.57e-02  &3.35e-05  &7.56e-07\\ \hline

\multirow{2}*{\boldsymbol{$F_{18}$}} 
&Mean    &4.02e+03  &1.49e+04  &4.85e+03  &\cellcolor{lightgray}3.67e+02\\ 
&Std     &1.16e+03  &6.65e+03  &2.63e+03  &2.20e+02\\ \hline

\multirow{2}*{\boldsymbol{$F_{19}$}} 
&Mean    &5.12e+05  &2.04e+05  &5.05e+05  &\cellcolor{lightgray}1.07e+04\\ 
&Std     &2.98e+04  &9.35e+03  &2.63e+04  &1.25e+03\\ \hline

\multirow{2}*{\boldsymbol{$F_{20}$}} 
&Mean    &1.13e+03  &1.20e+03  &1.19e+03  &\cellcolor{lightgray}7.97e-01\\ 
&Std     &1.93e+02  &1.55e+02  &1.07e+02  &1.64e+00\\ \hline
 \hline
 
\textbf{w-d-l}
&  &13-2-5  &13-4-3  &13-4-3  &-\\ \hline

\end{tabular}
\end{table}
\par}

\subsection{Algorithm Settings}
For the compared algorithms, three representatives of the existing DC-based EAs are tested, together with their parallelized counterparts. 
We denote these three DC-based EAs as DC-NG, DC-RG and DC-DG, and their parallelized counterparts as DC-NG-P, DC-RG-P and DC-DG-P. 
To be specific, DC-NG is short for divide-and-conquer with natural grouping, which naively decomposes the $D$-dimensional problem into $D$ 1-dimensional sub-problems. 
DC-RG follows the random grouping strategy proposed in \cite{yang2008large} that decomposes the decision variables into 10 sub-problems in an on-line manner, each of which exclusively consists of 100 decision variables. 
The on-line random grouping happens every iteration as \cite{omidvar2010more} proved that randomly grouping more frequently could minimize the interdependencies among sub-problems. 
DC-DG decomposes the problem using the improved differential grouping strategy, i.e., DG2 \cite{omidvar2017dg2}, which actively analyzes the interdependencies among decision variables so that the decomposition can be made much more accurate. 
The only difference between the above three compared algorithms and their corresponding parallelized counterparts is that the former builds the objective function for sub-problems using Eq.(2), while the latter builds that using Eq.(3). 
The decomposition phases of the above algorithms are parameterless.

To make the comparisons fair enough, all the compared algorithms employ the same sub-problem optimizer with NPDC. 
That is, in each sub-problem, an offspring is generated based on a parent using Eq.(10) and Eq.(11). 
The search step-size is adjusted with Eq.(12). 
For DC-NG and DC-NG-P, as the sub-problems are all 1-dimensional, each partial solution is actually a scalar, which is the same case with NPDC. 
For the rest compared algorithms, each partial solution is a vector, while the search step-size is still a scalar for each sub-problem. 
When applying Eq.(10) and Eq.(11) to a vector partial solution, the vector is actually generated one dimension at a time using the same search step-size, until all dimensions in the sub-problem have been gone through. 

In order to clearly see how the different ways of building objective functions for sub-problems will influence the performance of the tested algorithms, we let $\lambda=1$ for all tested algorithms.
As mentioned earlier, the CEC'2010 test suite involves two characteristics, i.e., multi-modality and separability. 
Between them, the different degrees of separability will largely influence the evaluation of partial solutions, which highly correlates with building $f^t_j$. 
Oppositely, the multi-modality mainly concerns how the sub-problem optimizers can explore the solution space more effectively, which has nothing to do with building $f^t_j$. 
Thus, it would be better not to enhance the exploration ability of the tested algorithms. 
Otherwise, it is difficult to tell whether a good qualitied solution is produced by effective objective functions for sub-problems or by an effective sub-problem optimizer. 
By letting $\lambda=1$, each sub-problem optimizer is merely an individual-based hill-climber, whose exploration ability has been minimized, and the impacts of building objective functions for sub-problems on the final solution quality can be emphasized.

\subsection{Investigations on the parallelism of existing DC-based EAs}
In this part, the above-mentioned three representatives of existing DC-based EAs, i.e., DC-NG, DC-RG, and DC-DG, are compared with their parallel counterparts, i.e., DC-NG-P, DC-RG-P, and DC-DG-P, respectively. 
The comparisons aim to show that, to parallelize existing DC-based EAs, the final solution quality may be degenerated. 
The empirical results on all 20 tested problems are shown in Figs.2(a)-(c), respectively for each pair of algorithms. 
In each figure, there are 20 bars representing the comparisons of each existing DC-based EAs against its parallel counterparts on 20 different tested problems. 
For each bar, the final solution qualities of a pair of two algorithms averaged over 20 problems are shown, where the blue bar represents the performance of the existing DC-based EA and the orange bar represents its parallel counterpart. 
As all the 20 problems are minimization problems, the shorter the bar is, the better the corresponding algorithm performs.

There are two observations can be made from Figs.2(a)-(c). 
First, for the three pairs of algorithms, the existing DC-based EAs all perform generally better than their parallel counterparts. 
On some problems, the superiority can be very significantly (see \#12 in all three figures for example), while on others it is only slightly better (see \#6 in Fig.2(c)). 
However, note that, sometimes 1\% better could be dominant. 
Second, the more accurate the decomposition is, the closer the performances between the existing algorithm and its parallel counterpart is. 
This is because the interdependencies among sub-problems will amplify the inconsistence of the objective functions between the sub-problems and the original problem \cite{panait2010theoretical}. 
That is, the higher the interdependencies exist between two sub-problems, their objective functions are more likely to be inconsistent with $f$.
Nevertheless, even the DG2 decomposition strategy has empirically shown the perfect decomposition, i.e., no interdependencies among sub-problems, on most problems in CEC'2010 test suite \cite{omidvar2017dg2}, the divergences between DC-DG and DC-DG-P on some problems are still significant. 
This phenomenon again stresses that the existing DC-based EAs are difficult to be parallelized, unless some compromises on the final solution qualities are made.

\subsection{Investigations on the effectiveness of NPDC}
Indeed, NPDC is proposed for its feature of being naturally parallelizable, while its effectiveness of reducing the search iterations or producing high-quality final solution is also important. 
Because if NPDC performs even worse than the parallelized compared algorithms, one should directly use them to solve problems rather than NPDC. 
In order to verify the effectiveness of NPDC, it is compared with DC-NG, DC-RG, and DC-DG on all 20 problems for 20 runs. 
The averaged final solution of each algorithm is shown in Table I, together with the standard deviation. 
For each problem, the best averaged solution quality is marked in gray for emphasis. 
The statistical test is given in the last row of Table I, where `w', `d', and `l' indicate the number of problems that NPDC performs statistically better, the same, and worse than the compared algorithm. 
Furthermore, the best intermediate solution qualities during the optimization of each algorithm are also recorded and averaged, which are depicted in Figs.3(a)-(t) to show the convergence rate of the compared algorithms. 
In these figures, the X-axis represents the number of function evaluations consumed by the algorithms, while the Y-axis describes the averaged solution quality at a $\log_{10}$ level. 
Hence, the Y-axis of the figures actually denotes the orders of magnitude of the final solution quality. 
On this basis, the lower the curve is, the faster the corresponding algorithm converges.

A quick conclusion can be drawn from Table I and Figs.3(a)-(t) that, NPDC successfully outperforms the three compared algorithms on the majority of 20 tested problems. 
Most importantly, even DC-DG has perfectly decomposed the sub-problems, it is still statistically inferior to NPDC, which just naively decomposes the problem into 1-dimensional sub-problems. 
This actually benefits from two aspects. 
First, NPDC evaluates the partial solutions consistent with the original problem, where a good partial solution to each sub-problem is definitely a good component to $f$. 
Second, existing DC-based EAs consume $M$ function evaluations in one iteration, each of which is used to evaluate the offspring partial solution in the $i$-th sub-problem using $f^t_i$, $i=1,2,...,M$. 
On the other hand, NPDC only consumes 1 function evaluation in one iteration, which gains $M$ times more iterations for optimization. 
Thus, NPDC is shown to be a competitive DC-based EA for large-scale optimization problems that can output very good-qualitied solution. 

\subsection{Investigations on the parallelism of NPDC}
To measure the parallelism of NPDC, we run NPDC on the 20 problems with 1, 2, 4, 6, 8 and 10 CPU cores, respectively. 
Three kinds of data are recorded for each run, i.e., the simulated computational time $\widehat{T_1}$ of NPDC on single-core, the simulated computational time $\widehat{T_N}$ of NPDC on $N$-cores, and the simulated computational time $\widehat{T_{FE}}$ for function evaluations. 
According to Eq.(13), we can calculate two kinds of speed-up ratios. 
The first is the theoretical speed-up ratio that all the components of NPDC, except the function evaluations, are assumed to be fully parallelized, calculated as $\frac{N}{1+\frac{\widehat{T_{FE}}}{\widehat{T_1}}\cdot(N-1)}$. 
The second is the simulated speed-up ratio that directly compares the simulated computational time of NPDC on single-core and $N$-cores, calculated as $\frac{\widehat{T_1}}{\widehat{T_N}}$. 
In general, the closer the latter to the former, the better the parallelism of NPDC will be. 
We only depict the results on the first 4 tested problems in Figs.4 (a)-(d) as illustrations.

It can be clearly seen that, the simulated speed-up of NPDC remains very close to its theoretical speed-up on the 4 problems. 
Actually, this also happens to the rest problems, which is omitted here due to the limitation of the pages. 
The results verify that NPDC is naturally parallelizable. 
We also noticed that for the 6-cores case, the gap between those two speed-up becomes a little bit larger. 
The reason might be that, for the tested problems, their 1000 decision variables cannot be equally assigned to 6 cores, where not all the current best partial solution can be collected simultaneously from each sub-problem for function evaluations. 
If we look at the Y-axis of the four figures, both speed-up ratios drop down dramatically on problems 2 and 3. 
This is because the computational costs for function evaluations of these two problems are relatively high, which are above 20\% of the total computational time as we counted. 
This corresponds to the discussions in section IV.C that the function evaluation phase is the main bottleneck for the parallel efficiency of NPDC. 
Unfortunately, such bottleneck exists for all DC-based EAs due to the black-box optimization nature, and thus cannot be addressed systematically.

\subsection{Investigations on the meta-model of NPDC}
The meta-model plays an important role to the performance of NPDC. 
In order to show how the accuracy of the meta-model can benefit NPDC, we specially designed a new compared algorithm, named NPDC-random. 
In NPDC-random, the two parameters of the meta-model, i.e., $PL^t_j$ and $PS^t_j$, are fixed to 0.5 during the whole optimization. 
This means that, given an offspring and its parent, the meta-model in NPDC-random simply gives a random guess on which one is better. 
Except for this, the rest sub-routines of NPDC-random are kept the same to NPDC. 
Both algorithms are compared on the 20 problems and the averaged final solution qualities over 20 runs are shown in Fig.5. 
On problems 1, 4, 7, 8, 9, 12, 13, 14, 17, 18, 19, 20, NPDC outperforms significantly than NPDC-random. For the rest problems, NPDC only shows slightly advantages over NPDC-random. 
The reason might be that the landscapes of those problem are not smooth at all, where the assumption of the proposed meta-model does not hold anymore and the meta-model becomes less effective.

\section{Conclusions and Discussions}
This paper investigated the parallelism of DC-based EAs on large-scale optimization black-box problems. 
We first discussed that the existing DC-based EAs could not be directly parallelized. 
Otherwise, their effectiveness in terms of final solution quality could be severely degenerated. 
The reason behind this phenomenon was also analyzed in details. 
To be specific, one most important thing for DC-based EAs is to build the objective functions for sub-problems. 
In existing works, the objective functions for sub-problems were inconsistent with the objective function for the original problem. 
In other words, a partial solution that is good to a sub-problem may not be a component of a good complete solution to the original problem. 
We revealed an ideal way that could build objective functions for sub-problems consistent with the original problem. 
Then the ways for building objective functions used by existing DC-based EAs were compared to the ideal way. 
It shows that the parallelizable way of existing DC-based EAs can be probabilistically much worse than the non-parallelizable way. 
This analysis was also supported by empirical studies. 
We then designed a novel way based on the ideal way, which is naturally parallelizable and effective. 
The resultant algorithm, i.e., NPDC, was detailed. 
Its effectiveness in terms of the final solution quality was empirically verified on the CEC'2010 large-scale optimization benchmark, against several representatives of existing DC-based EAs. 
We also empirically showed that NPDC can be efficiently parallelized. 
Besides, the limitations of both NPDC and its meta-model are also discussed in this paper.

It should be noted that the framework of NPDC looks similar to our previous work \cite{yang2018turning} that also divides the original problem into $D$ sub-problems and employs an meta-model in each sub-problem.
However, this work is actually very different from the work \cite{yang2018turning} for two reasons.
First, these two works have different motivations and contributions. 
For \cite{yang2018turning}, it aims to solve the difficulty of accurately complementing a partial solution, which is in turn addressed as an expensive optimization problem and the meta-model is thus used for cheaper computational costs. 
However, for this work, it aims to address the parallelism difficulty of existing DC-based EAs, where the meta-model is used to break the dining phylosophers dilemma.
From this perspective, these two works are proposed for completely different problems but share the same framework.
Second, from the perspective of implementations under the framework, these two algorithms are still very different.
For \cite{yang2018turning}, each sub-problem employs a (1+$\lambda$)-EA paradigm where one parent generates $\lambda$ offsprings.
Among them, half of the $\lambda$ offsprings are generated with the Gaussian mutation operator, and half of them are generated with the Cauchy mutation operator.
At the end of each iteration, the parent will be compared with each offspring one-by-one for survival.
However, for this work, each sub-problem only employs a (1+1)-EA, making the algorithm much simpler and more compatible with other well-established EAs, e.g., PSO and DE.
On this basis, these two works are inherently different from each other, and thus we did not discuss the details of \cite{yang2018turning} in the previous sections.

In NPDC, the sub-problem optimizer can be a population-based EA following the (1+1)-EA paradigm. 
Specially, in this paper, we let it be $\lambda$ independent individual-based hill-climbers. 
In the future, it is expected that the NPDC could be incorporated with other well-established (1+1)-EA paradigm based EAs, e.g., DE and PSO, to enhance its ability of solving complex multi-modal optimization problems, while maintaining the feature of natural parallelism to a good extent. 
Besides, NPDC currently employs the Gaussian mutation and Cauchy mutation operators in the sub-problem optimizers, making it ill-equipped to deal with combinatorial optimization problems. 
This issue will also be investigated for future work.

\ifCLASSOPTIONcompsoc
  \section*{Acknowledgments}
\else
  \section*{Acknowledgment}
\fi

This work was supported by National Key R\&D Program of China (Grant No. 2017YFC0804002), the Natural Science Foundation of China (61806090 and 61672478), the Program for Guangdong Introducing Innovative and Enterpreneurial Teams (Grant No. 2017ZT07X386), Shenzhen Peacock Plan (Grant No. KQTD2016112514355531), and the Program for University Key Laboratory of Guangdong Province (Grant No. 2017KSYS008).

\ifCLASSOPTIONcaptionsoff
  \newpage
\fi

\bibliographystyle{IEEEtran}
\bibliography{IEEEabrv,NPDCrefer}

\end{document}